\documentclass[11pt]{article}

\usepackage{acl}

\usepackage{latexsym}

\usepackage[T1]{fontenc}
\usepackage[utf8]{inputenc}

\usepackage{microtype}

\usepackage{inconsolata}

\usepackage{graphicx}
\usepackage{booktabs}
\usepackage{multirow}
\usepackage{tabularx}
\usepackage{enumitem}

\usepackage{tikz}
\usetikzlibrary{positioning,calc,arrows}
\usepackage{pgfplots}
\pgfplotsset{compat=1.18}

\title{CSCBench: A PVC Diagnostic Benchmark for Commodity Supply Chain Reasoning}

\author{
    \begin{tabular}{c}
    Yaxin Cui\textsuperscript{1,2}\thanks{Corresponding author.},
    Yuanqiang Zeng\textsuperscript{1},
    Jiapeng Yan\textsuperscript{2},
    Keling Lin\textsuperscript{3},
    Kai Ji\textsuperscript{4} \\
    Jianhui Zeng\textsuperscript{2},
    Sheng Zhang\textsuperscript{2},
    Xin Luo\textsuperscript{2},
    Binzhu Su\textsuperscript{1},
    Chaolai Shen\textsuperscript{1}\thanks{Corresponding author.},
    Jiahao Yu\textsuperscript{1}\thanks{Corresponding author.},
    \end{tabular}
    \\[0.75ex]
    \begin{tabular}{c}
    \textsuperscript{1}Xiamen SmartChain Innovations Co., Ltd., Xiamen, China \\
    \textsuperscript{2}Xiamen ITG Digital Technology Co., Ltd., Xiamen, China \\
    \textsuperscript{3}Xiamen C\&D Co., Ltd., Xiamen, China \\
    \textsuperscript{4}Xiamen Xiangyu Co., Ltd., Xiamen, China
    \end{tabular}
    \\[0.5ex]
    {\footnotesize
    \begin{tabular}{c}
    \texttt{cui\_yaxin@outlook.com, linkl@chinacdc.com, xyjik@xiangyu.cn} \\
    \texttt{\{zengyq, subz, yujh, zengjh, zhangsheng, yanjp, luoxin, shencl\}@xindeco.com.cn}
    \end{tabular}
    }
}

\begin{document}
\maketitle
\begin{abstract}
Large Language Models (LLMs) have achieved remarkable success in general benchmarks, yet their competence in commodity supply chains (CSCs)---a domain governed by institutional rule systems and feasibility constraints---remains under-explored. CSC decisions are shaped jointly by process stages (e.g., planning, procurement, delivery), variety-specific rules (e.g., contract specifications and delivery grades), and reasoning depth (from retrieval to multi-step analysis and decision selection). We introduce \textbf{CSCBench}, a 2.3K+ single-choice benchmark for CSC reasoning, instantiated through our PVC 3D Evaluation Framework (Process, Variety, and Cognition). The Process axis aligns tasks with SCOR+Enable \citep{noauthor_scor_2010}; the Variety axis operationalizes commodity-specific rule systems under coupled \emph{material--information--financial} constraints \citep{cooper_supply_1997}, grounded in authoritative exchange guidebooks/rulebooks and industry reports; and the Cognition axis follows Bloom's revised taxonomy \citep{anderson_taxonomy_2001}. Evaluating representative LLMs under a direct prompting setting, we observe strong performance on the Process and Cognition axes but substantial degradation on the Variety axis, especially on Freight Agreements. CSCBench provides a diagnostic yardstick for measuring and improving LLM capabilities in this high-stakes domain.
\end{abstract}

\section{Introduction}

Commodity supply chains (CSCs) move trillions of dollars of energy, metals, and agricultural products each year and underpin many downstream industries. Unlike many consumer-product supply chains, CSC decisions are jointly shaped by institutional and compliance rules and physical feasibility constraints (e.g., contract specifications and delivery grades, port/vessel constraints, and time-sensitive logistics), and often couple physical operations with financial pricing and risk management. In practice, answering seemingly simple questions---e.g., identifying quality penalties in a delivery specification, determining whether a cargo is deliverable under a given contract, which shipping route is feasible under vessel/port constraints, or how a disruption shifts sourcing choices---requires reading and reasoning over technical rule texts. As Large Language Models (LLMs) become decision-support tools, reliably measuring their competence in this domain and localizing failure modes is increasingly important.

Despite rapid progress in LLM benchmarking, there is no rigorous, diagnostic evaluation standard tailored to commodity supply chains. Existing benchmarks generally fall into three categories, none of which adequately cover CSC reasoning under institutional rules and feasibility constraints:
\begin{enumerate}[leftmargin=*,noitemsep]
  \item \textbf{General knowledge benchmarks} (e.g., MMLU \citep{hendrycks_measuring_2020}, C-Eval \citep{huang_c-eval_2023}) are broad, but are not designed to diagnose process-grounded competence (e.g., SCOR+Enable stages) or rule-intensive, variety-specific decision constraints.
  \item \textbf{Financial benchmarks} (e.g., FinEval \citep{guo_fineval_2025}, BizBench \citep{krumdick_bizbench_2024}) probe financial literacy, yet typically abstract away operational feasibility and contractual rule texts (contract specs, delivery grades, shipping terms) that govern commodity operations.
  \item \textbf{Operations research datasets} excel at optimization over structured inputs, but rarely evaluate language-based interpretation of real-world rules and narratives or provide diagnostic slices aligned with industrial workflows.
\end{enumerate}

To bridge this gap, we introduce \textbf{CSCBench}, a benchmark dedicated to commodity supply chain reasoning. CSCBench is designed as a diagnostic framework instantiated through our PVC (Process--Variety--Cognition) 3D Evaluation Framework: (i) Process aligns tasks to an end-to-end operational lifecycle (SCOR+Enable \citep{noauthor_scor_2010}); (ii) Variety evaluates commodity-specific institutional and contractual constraints derived from authoritative documents (e.g., exchange guidebooks, contract specifications, and industry reports); and (iii) Cognition stratifies required reasoning depth following Bloom's revised taxonomy \citep{anderson_taxonomy_2001}. Using a unified single-choice format, CSCBench supports interpretable slice-based reporting rather than a single opaque score.

\begin{table*}[t]
  \centering
  \scriptsize
  \setlength{\tabcolsep}{2pt}
  \renewcommand{\arraystretch}{1.1}
  \begin{tabular}{l | c ccc | c ccc | c ccccc}
  \toprule
  \multirow{2}{*}{\textbf{Model}} & \multicolumn{4}{c|}{\textbf{Process (X)}} & \multicolumn{4}{c|}{\textbf{Variety (Y)}} & \multicolumn{6}{c}{\textbf{Cognition (Z)}} \\
   & \textbf{Avg} & CIPS & CSCP & SCMP & \textbf{Avg} & Soy & Iron & Frt. & \textbf{Avg} & Com.Trd & SCM & Int.Trd & Fut/Opt & Logist. \\
  \midrule
  deepseek & \textbf{86.1} & 94.3 & 77.4 & 88.7 & \textbf{54.5} & 71.4 & 63.1 & 28.9 & \textbf{95.7} & 98.9 & 94.3 & 94.1 & 93.3 & 97.7 \\
  gemini & \textbf{91.6} & 95.3 & 87.0 & 93.4 & \textbf{62.0} & 68.9 & 68.8 & 48.2 & \textbf{96.3} & 98.7 & 93.6 & 95.7 & 95.5 & 98.1 \\
  glm & \textbf{88.0} & 95.2 & 77.1 & 91.7 & \textbf{61.4} & 74.6 & 64.4 & 45.2 & \textbf{96.8} & 99.0 & 94.8 & 96.0 & 95.9 & 98.3 \\
  qwen & \textbf{84.4} & 93.2 & 72.7 & 87.2 & \textbf{57.6} & 73.7 & 67.5 & 31.6 & \textbf{94.4} & 99.9 & 88.6 & 94.7 & 92.3 & 96.8 \\
  \bottomrule
  \end{tabular}
  \caption{Detailed performance breakdown. Axis scores (Avg) are the macro-average of their respective sub-benchmarks (X:3, Y:3, Z:5). Runs with credit-related API failures (e.g., OpenRouter 402) are excluded from aggregation. -- indicates no valid run data.}
  \label{tab:axis_results_outputs}
\end{table*}

\paragraph{Main finding.}
Evaluating representative frontier and open-source LLMs under a direct prompting setting, we find that models perform strongly on the Process and Cognition slices, but degrade substantially on the Variety axis. In particular, Freight Agreements is consistently the hardest sub-benchmark. This suggests that commodity-specific institutional rules---rather than generic language fluency---remain a primary bottleneck for practical deployment.

Our key contributions are:
\begin{itemize}[leftmargin=*,noitemsep]
  \item \textbf{PVC diagnostic framework}: We decompose CSC competence into Process (SCOR+Enable-aligned lifecycle \citep{noauthor_scor_2010}), Variety (commodity-specific institutional rule systems under coupled material--information--financial constraints \citep{cooper_supply_1997}), and Cognition (Bloom-style depth \citep{anderson_taxonomy_2001}).
  \item \textbf{A native-source benchmark}: CSCBench contains 2,300+ single-choice questions curated from authoritative exchange guidebooks/rulebooks, professional exam materials, and industry reports, enabling objective scoring under a unified schema.
  \item \textbf{PVC-sliced analysis}: We report stage-, variety-, and cognition-level breakdowns, exposing that commodity-specific institutional constraints---rather than generic language fluency---remain a primary barrier for practical deployment.
\end{itemize}

% \paragraph{Paper organization.}
% Section~\ref{sec:related_work} reviews related benchmarks. Section~\ref{sec:pvc_framework} introduces the PVC framework. Section~\ref{sec:data_pipeline} describes dataset construction. Sections~\ref{sec:experimental_setup} and \ref{sec:results} present the evaluation protocol and results.

\section{Related Work}

\subsection{General and Domain-Specific LLM Evaluation}

General benchmarks and evaluation suites such as MMLU \citep{hendrycks_measuring_2020}, C-Eval \citep{huang_c-eval_2023}, HELM \citep{liang2023holisticevaluationlanguagemodels}, and BIG-bench \citep{srivastava2023imitationgamequantifyingextrapolating} are widely used to track broad world knowledge and reasoning across academic subjects. Complementary platforms based on human preference or model-based judging (e.g., Chatbot Arena and MT-Bench \citep{chiang2024chatbotarenaopenplatform,zheng2023judgingllmasajudgemtbenchchatbot}) have further popularized interactive evaluations. While indispensable, these benchmarks are primarily subject-centric and do not align evaluation with end-to-end operational lifecycles (e.g., SCOR \citep{noauthor_scor_2010}) or report diagnostics that reflect real industrial workflows. As a result, strong performance on these suites provides limited evidence that a model can support process-grounded decisions in complex operational domains such as supply chains.

Specialized benchmarks assess LLM competence under professional conventions in high-stakes domains. For example, MedBench \citep{cai_medbench_2024} and PubMedQA \citep{jin_pubmedqa_2019} evaluate biomedical knowledge and clinical-style QA; LawBench \citep{fei_lawbench_2024}, LegalBench \citep{guha_legalbench_2023}, and legal/contract benchmarks such as LexGLUE, ContractNLI, and CUAD \citep{chalkidis-etal-2022-lexglue,koreeda2021contractnlidatasetdocumentlevelnatural,hendrycks2021cuadexpertannotatednlpdataset} test legal language understanding and contractual reasoning; SciBench \citep{wang_scibench_2024} targets college-level scientific problem solving. Financial benchmarks such as FinEval \citep{guo_fineval_2025}, BizBench \citep{krumdick_bizbench_2024}, and FinQA \citep{chen2022finqadatasetnumericalreasoning} probe financial knowledge and quantitative reasoning. These efforts provide depth within their respective ontologies, but they do not treat commodity-specific institutional rule systems (e.g., exchange contract specifications, trading guides) and operational feasibility constraints (e.g., grades, delivery timing, logistics capacity) as first-class evaluation targets.

\subsection{Supply Chain Evaluation and Constraint-Grounded Reasoning}

Supply chain research traditionally evaluates decision-making via optimization and quantitative models \citep{simchi-levi_managing_2004}, often over structured numerical inputs (demands, costs, capacities), which does not capture a model's ability to interpret natural-language market narratives, contracts, and exceptions. Conversely, generic reasoning benchmarks such as GSM8K \citep{cobbe_training_2021}, BIG-Bench Hard \citep{suzgun_challenging_2023}, and StrategyQA \citep{geva_strategyqa_2021} probe multi-step reasoning but are largely constraint-agnostic. Commodity supply chains amplify this gap because decisions are governed by explicit institutional rules and hard feasibility constraints; a common failure mode is producing fluent yet infeasible answers. CSCBench addresses this gap by providing a PVC-decomposed evaluation that localizes failures by \emph{process stage}, \emph{commodity variety}, and \emph{cognitive depth}.

\subsection{Supply Chain Theory Foundations and Commodity Constraints}

Classic supply chain theory characterizes \emph{executable} decisions as the joint outcome of multiple constraints: process stages determine where a decision sits and what actions are available (e.g., SCOR, including Enable \citep{noauthor_scor_2010}); multi-actor coordination introduces information distortion and incentive misalignment that can manifest as systematic amplification and coordination failures (e.g., the bullwhip effect and contract-based coordination \citep{lee_bullwhip_1997,cachon_revenuesharing_2005}); and exogenous disruptions can propagate through coupled networks and generate persistent performance and risk consequences \citep{hendricks_glitches_2005}. Commodity supply chains make these constraints especially explicit via institutional rule texts (contract specifications, delivery/grade rules, penalties) and hard feasibility limits (ports/vessels/routes/timing), often coupled with pricing, hedging, and settlement mechanisms. CSCBench operationalizes these concerns into a scored PVC diagnosis: \emph{Process} anchors tasks to SCOR+Enable stages, \emph{Variety} captures commodity-specific rule systems and feasibility constraints, and \emph{Cognition} stratifies the required reasoning depth.

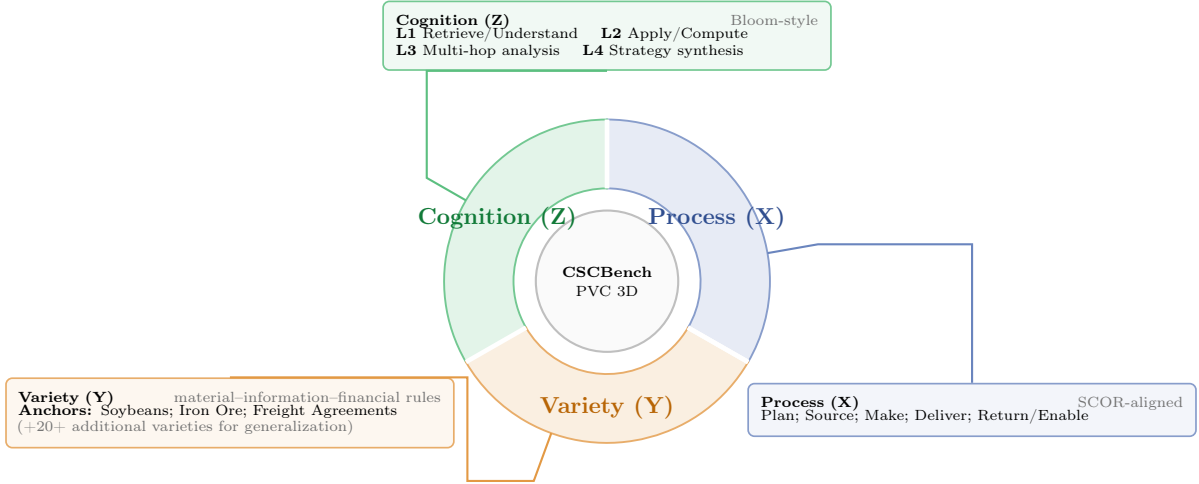
\begin{figure*}[t]
  \centering
  % PVC framework overview (taxonomy-style: center + colored ring sectors + callouts).
  \resizebox{0.98\textwidth}{!}{%
  \begin{tikzpicture}[
    font=\scriptsize,
    >=latex,
    callout/.style={rounded corners=3pt, thick, inner sep=6pt, align=left, text width=7.0cm},
  ]
    % palette
    \definecolor{pvcBlue}{RGB}{56,92,168}
    \definecolor{pvcOrange}{RGB}{217,119,6}
    \definecolor{pvcGreen}{RGB}{22,163,74}

    % radii
    \def\rC{1.18}   % center radius
    \def\rI{1.55}   % inner ring radius
    \def\rO{2.70}   % outer ring radius

    % center
    \filldraw[fill=black!2, draw=black!25, line width=1.0pt] (0,0) circle (\rC);
    % center label (avoid overlap by using positive line spacing)
    \node[align=center] at (0,0) {\bfseries CSCBench\\[1pt]\scriptsize PVC 3D};

    % ring sectors (120 degrees each)
    % Process (X): -30 to 90
    \path[fill=pvcBlue!12, draw=pvcBlue!60, line width=0.9pt]
      (-30:\rI) arc (-30:90:\rI) -- (90:\rO) arc (90:-30:\rO) -- cycle;
    % Cognition (Z): 90 to 210
    \path[fill=pvcGreen!12, draw=pvcGreen!60, line width=0.9pt]
      (90:\rI) arc (90:210:\rI) -- (210:\rO) arc (210:90:\rO) -- cycle;
    % Variety (Y): 210 to 330
    \path[fill=pvcOrange!12, draw=pvcOrange!60, line width=0.9pt]
      (210:\rI) arc (210:330:\rI) -- (330:\rO) arc (330:210:\rO) -- cycle;

    % separators between sectors (improve visual grouping)
    \draw[white, line width=2.2pt] (-30:\rI) -- (-30:\rO);
    \draw[white, line width=2.2pt] (90:\rI) -- (90:\rO);
    \draw[white, line width=2.2pt] (210:\rI) -- (210:\rO);
    \draw[white, line width=2.2pt] (330:\rI) -- (330:\rO);

    % labels on ring
    \node[font=\bfseries, text=pvcBlue!85!black] at (30:2.10) {Process (X)};
    \node[font=\bfseries, text=pvcGreen!75!black] at (150:2.10) {Cognition (Z)};
    \node[font=\bfseries, text=pvcOrange!85!black] at (270:2.10) {Variety (Y)};

    % callout boxes
    \node[callout, draw=pvcGreen!60, fill=pvcGreen!6] (zbox) at (0,4.10) {%
      \textbf{Cognition (Z)} \hfill \textcolor{black!55}{Bloom-style}\\[-2pt]
      \textbf{L1} Retrieve/Understand \quad
      \textbf{L2} Apply/Compute\\
      \textbf{L3} Multi-hop analysis \quad
      \textbf{L4} Strategy synthesis
    };

    \node[callout, draw=pvcOrange!60, fill=pvcOrange!6] (ybox) at (-6.25,-2.20) {%
      \textbf{Variety (Y)} \hfill \textcolor{black!55}{material--information--financial rules}\\[-2pt]
      \textbf{Anchors:} Soybeans; Iron Ore; Freight Agreements\\
      \textcolor{black!55}{(+20+ additional varieties for generalization)}
    };

    \node[callout, draw=pvcBlue!60, fill=pvcBlue!6] (xbox) at (6.05,-2.15) {%
      \textbf{Process (X)} \hfill \textcolor{black!55}{SCOR-aligned}\\[-2pt]
      Plan; Source; Make; Deliver; Return/Enable
    };

    % connectors (clean radial + orthogonal routing to avoid "loops")
    \draw[pvcGreen!70, line width=1.0pt]
      (150:\rO) -- ++(150:0.75) |- (zbox.south);
    % route Y-connector around the callout box to avoid overlap
    \draw[pvcOrange!75, line width=1.0pt]
      (250:\rO) -- ++(250:0.85) -- ++(-1.10,0) |- (ybox.north);
    \draw[pvcBlue!75, line width=1.0pt]
      (10:\rO) -- ++(10:0.85) -| (xbox.north);
  \end{tikzpicture}%
  }
  \caption{Overview of the proposed PVC 3D Evaluation Framework for commodity supply chains. The framework decomposes evaluation along three orthogonal axes: \textbf{Process} (SCOR-aligned stages), \textbf{Variety} (commodity-specific rule systems and anchors), and \textbf{Cognition} (Bloom-style reasoning depth). Axis scores are macro-averages of their sub-benchmarks.}
  \label{fig:pvc_framework}
\end{figure*}

\section{The PVC 3D Evaluation Framework}

To evaluate LLMs in commodity supply chains---a domain with coupled \emph{physical} and \emph{financial} constraints---we propose the \textbf{PVC (Process--Variety--Cognition) 3D Evaluation Framework}. The three axes are designed to be largely orthogonal and jointly diagnostic: the \textbf{Process (X)} axis asks \emph{where} a task sits in the end-to-end workflow; the \textbf{Variety (Y)} axis asks \emph{which} commodity universe and rule system it belongs to; and the \textbf{Cognition (Z)} axis asks \emph{how deeply} the model must reason to answer it. This decomposition supports interpretable reporting: a strong overall score can still hide bottlenecks concentrated in specific process stages, varieties, or reasoning depths.

\paragraph{Why three dimensions?}
In practice, failures in commodity supply chain reasoning typically stem from three non-substitutable sources: (i) \emph{process misalignment} (reasoning under the wrong SCOR stage), (ii) \emph{variety/rule mismatch} (confusing commodity-specific terminology and rule systems), and (iii) \emph{insufficient cognitive depth} (stopping at shallow retrieval or template application instead of multi-step reasoning and synthesis). PVC explicitly separates these factors to enable actionable diagnosis rather than a single opaque score.

\subsection{X-Axis: SCOR+Enable-Aligned Process}
For the X-axis, we adopt \textbf{SCOR+Enable} \citep{noauthor_scor_2010} as an explicit process backbone and align questions to end-to-end process stages (Plan/Source/Make/Deliver/Return/Enable), enabling process-level diagnosis. We instantiate this axis with three professional qualification sub-benchmarks (CIPS/CSCP/SCMP) that provide broad coverage of supply-chain decision-making along the workflow.

\subsection{Y-Axis: Variety as Rule- and Constraint-Consistency}
Unlike the Process axis, the Y-axis focuses on \textbf{variety-specific constraint consistency} in a given commodity context: can a model make an executable judgment by adhering to institutional rule texts and feasibility conditions (e.g., contract specifications, delivery/grade clauses, and freight/settlement terms), rather than relying on generic knowledge alone? We operationalize this axis with a five-layer ladder (L1--L5) that progresses from basic definitions to more complex market mechanisms and business scenarios. The current release anchors on three deep-dive varieties---\textbf{Iron Ore}, \textbf{Soybeans}, and \textbf{Freight Agreements}---curated from publicly verifiable sources such as exchange rulebooks/guides, contract and delivery specifications, methodology documents, and industry reports.

\subsection{Z-Axis: Bloom-Grounded Cognitive Operations}
The Z-axis characterizes the depth of \emph{cognitive operations} required by a question. We stratify items using Bloom's revised taxonomy \citep{anderson_taxonomy_2001} (e.g., understanding concepts/rules, applying calculations, and analyzing scenarios under constraints), while keeping the same objective single-choice format across all axes. We instantiate Z with five course-style sub-benchmarks: \textbf{Supply Chain Management} (SCM; inventory/planning and operations basics), \textbf{Logistics} (transportation and warehousing basics), \textbf{International Trade} (trade terms and settlement basics), \textbf{Commodity Trade} (spot mechanisms and trading practice), and \textbf{Futures \& Options} (derivatives and hedging basics).

\begin{figure*}[t]
  \centering
  \includegraphics[width=\textwidth]{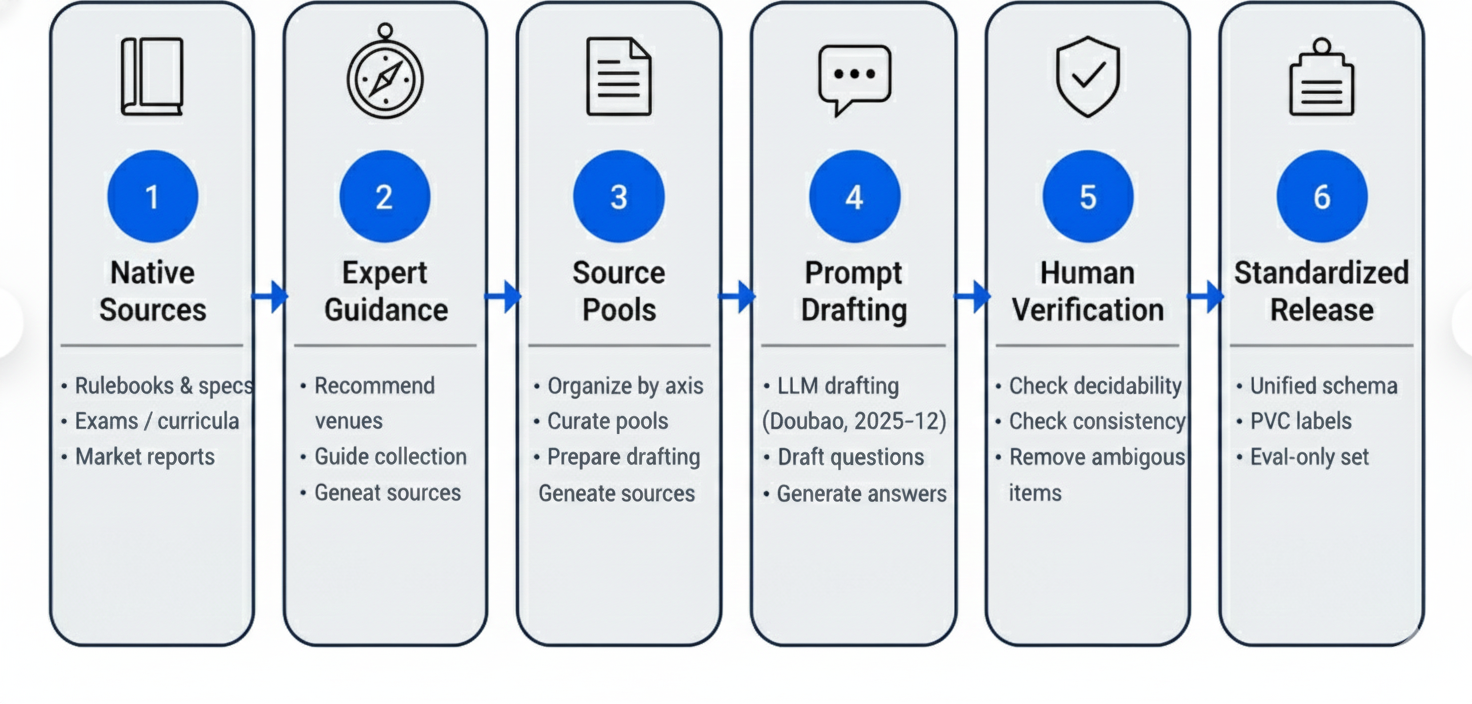}
  \caption{Data construction pipeline for CSCBench (high-level). Source discovery and collection are informed by domain experts; we curate the collected materials into axis-aligned source pools, draft items with an LLM under prompts, and apply human verification for decidability and internal consistency under a unified single-choice schema.}
  \label{fig:pipeline}
\end{figure*}

\section{Data Construction Pipeline}

To support \textbf{domain validity}, \textbf{auditability}, and \textbf{standardized evaluation}, we construct CSCBench with a \textbf{native-source, prompt-driven, human-verified} pipeline. The current release is \textbf{evaluation-only}: CSCBench is provided as a single test set.

\paragraph{Design rationale.}
Guided by domain experts, we discover and collect domain-authentic source materials and curate them into axis-aligned source pools to keep items verifiable against written evidence. We adopt an objective single-choice format to enable large-scale, reproducible scoring and fine-grained PVC-sliced analysis. Prompt-driven drafting improves coverage and throughput, while human verification screens for decidability and internal consistency.

\paragraph{Adversarial option perturbation.}
To reduce shortcut cues in multiple-choice evaluation and make the \textbf{Variety (Y)} axis more discriminative, we apply an \textbf{adversarial option perturbation} procedure during item construction: given a verified correct option, we rewrite distractors to be \emph{semantically closer} to the correct option while remaining inconsistent with the governing rule texts (e.g., contract clauses, grade specifications, freight/settlement terms). Domain experts then re-check that each item has a single unambiguous correct answer under the curated sources. This perturbation is used for the Y-axis varieties in the current release, and is not applied to the X/Z axes.

\subsection{Source Curation}
Source discovery and collection are informed by domain experts. We curate three source categories aligned with the PVC framework and instantiate each axis with three sub-benchmarks selected for complementary coverage rather than exhaustive representation:
\begin{itemize}[leftmargin=*,noitemsep]
  \item \textbf{X-axis (Process; SCOR+Enable coverage)}: professional supply-chain qualification materials (e.g., exams and syllabi), instantiated as CIPS/CSCP/SCMP.
  \item \textbf{Y-axis (Variety; five-layer ladder grounded in material, information, and financial flows)}: public, variety-specific documents (e.g., exchange rulebooks, contract specifications, and methodology documents), anchored on iron ore, soybeans, and dry-bulk freight agreements.
  \item \textbf{Z-axis (Cognition; Bloom-based depth)}: foundational materials spanning supply chain/logistics, international trade, and derivatives.
\end{itemize}

\subsection{Item Schema and Labels}
Each example follows a unified single-choice schema (question, options, answer, and an explanation) and is assigned a primary PVC label for slice-based reporting. The \texttt{explanation} field provides a source-grounded justification consistent with the selected answer for internal auditing; we maintain internal traceability to sources but do not release per-item source locations.

\subsection{Prompt-Driven Drafting and Human Verification}
We draft questions in a prompt-driven manner using a large language model (Doubao, December 2025 version), followed by human verification. Human verification screens for (i) decidability under the curated sources, (ii) internal consistency among question, options, answer, and explanation, and (iii) ambiguity (e.g., multiple plausible correct options) when identified. Items that fail these checks are excluded.

\subsection{Dataset Statistics}
The current version of CSCBench comprises \textbf{2,300+} multiple-choice questions, organized by the \textbf{3 axes $\times$ 3 sub-benchmarks} structure. The distribution is as follows:
\begin{itemize}[leftmargin=*,noitemsep]
  \item \textbf{X-Axis (Process)}: Includes questions from three major professional certifications (CIPS, CSCP, SCMP), comprehensively covering the six SCOR process stages: Plan, Source, Make, Deliver, Return, and Enable.
  \item \textbf{Y-Axis (Variety)}: Covers three anchor commodities: Iron Ore, Soybeans, and Freight Agreements. Each variety includes complete data across the \textbf{L1--L5} ladder, ranging from basic definitions to enterprise-level risk management.
  \item \textbf{Z-Axis (Cognition)}: Encompasses three foundational disciplines: Supply Chain Management, International Trade, and Derivatives, stratified by Bloom's cognitive depth.
\end{itemize}

We record generation and verification metadata for auditing; we do not remove near-duplicates, double-annotate labels, or control for potential pretraining overlap. Detailed statistics are presented in Table~\ref{tab:dataset_stats}.

\begin{table}[t]
    \centering
    \small
    \setlength{\tabcolsep}{4pt}
    \renewcommand{\arraystretch}{1.05}
    \begin{tabularx}{\columnwidth}{l c X}
    \toprule
    \textbf{Sub-benchmark} & \textbf{Count} & \textbf{Coverage} \\
    \midrule
    \multicolumn{3}{l}{\textbf{X (Process)}} \\
    CIPS & 311 & Focus: Source, Enable \\
    CSCP & 146 & End-to-End (Plan--Return) \\
    SCMP & 598 & Modular (Plan, Source, Make, Deliver) \\
    \midrule
    \multicolumn{3}{l}{\textbf{Y (Variety)}} \\
    Iron Ore & 160 & L1--L5 (Full Ladder) \\
    Soybeans & 147 & L1--L5 (Full Ladder) \\
    Freight Agreements & 223 & L1--L5 (Full Ladder) \\
    \midrule
    \multicolumn{3}{l}{\textbf{Z (Cognition)}} \\
    SCM \& Logistics & 304 & Bloom L1--L4 \\
    Trade Fundamentals & 303 & Bloom L1--L4 \\
    Derivatives & 150 & Bloom L1--L4 \\
    \midrule
    \textbf{Total} & \textbf{2,342} & \textbf{3 Axes $\times$ 3 Sub-benchmarks} \\
    \bottomrule
    \end{tabularx}
    \caption{Detailed statistics of the CSCBench dataset, organized by the PVC framework.}
    \label{tab:dataset_stats}
\end{table}

\section{Experimental Setup}
We now describe how we instantiate CSCBench to evaluate existing LLMs. Our goals are (i) to provide a fair comparison across model families, and (ii) to expose where along the PVC dimensions current systems fail.

\subsection{Research Questions}

We organize our experiments around three guiding questions:
\begin{itemize}[leftmargin=*,noitemsep]
  \item \textbf{RQ1 (overall competence)}: How well do mainstream LLMs perform on CSCBench overall?
  \item \textbf{RQ2 (PVC decomposition)}: Along which Process stages, Variety anchors, and Cognition levels do models succeed or fail?
  \item \textbf{RQ3 (CSC insight)}: Do model failures concentrate on CSC-specific executable reasoning under institutional rules and physical feasibility constraints, especially where these couple with pricing and risk-management mechanisms?
\end{itemize}

\subsection{Models}

We evaluate four representative LLMs (qwen, gemini, glm, deepseek). All models are evaluated \textbf{inference-only} on CSCBench; no additional fine-tuning is performed on our benchmark.

\subsection{Evaluation Protocol}
We use a C-Eval-style single-choice evaluation harness with direct prompting only. Each item is presented with its question and options, and the model is instructed to reply with \emph{only} the option letter. We apply a robust post-processing rule to extract the final letter (preferring explicit ``Answer:'' patterns when present, otherwise taking the last standalone option letter); non-parsable outputs are treated as incorrect.

We set temperature to 0.1 and use a fixed maximum output length for generation, and repeat each (sub-benchmark, model) evaluation for 5 runs under identical settings. Empty responses are not counted toward the denominator by default.

\subsection{Metrics and Axis-Level Reporting}
We report accuracy for single-choice questions. Let $C$ be the confusion matrix over answer choices; accuracy is computed as:
\[
\mathrm{Acc}=\frac{\sum_k C_{kk}}{\sum_{i,j} C_{ij}}.
\]
This reduces to $\mathrm{Acc}=\frac{TP+TN}{TP+TN+FP+FN}$ in the binary case. Empty responses are excluded from the denominator by default; non-parsable outputs are treated as incorrect.
For axis-level reporting, we macro-average across sub-benchmarks within each axis (equal weight) to avoid dominance by larger sub-benchmarks. Results are reported both globally and by axis/sub-benchmark.

\section{Results and Analysis}
\label{sec:results}
In this section, we report empirical results on CSCBench and analyze them along the PVC dimensions. Unless otherwise noted, all numbers refer to the benchmark (evaluation-only set) computed from our evaluation outputs. Runs with credit-related API failures are excluded from aggregation (Table~\ref{tab:axis_results_outputs}).

\subsection{Overall Performance}

Table~\ref{tab:axis_results_outputs} summarizes performance by axis and sub-benchmark, computed from evaluation outputs. Axis scores (Avg) are the macro-average of their respective sub-benchmarks; missing entries indicate that no valid run data is available.

\paragraph{Main finding.}
Across the evaluated models, accuracy is consistently high on the \textbf{Process (X)} and \textbf{Cognition (Z)} axes, but substantially lower on the \textbf{Variety (Y)} axis. In particular, \textbf{Freight Agreements (Frt.)} is the hardest sub-benchmark (28.9--48.2 in Table~\ref{tab:axis_results_outputs}), indicating that rule- and contract-intensive, constraint-coupled questions remain challenging even when models perform strongly on other slices. We also observe a small number of \emph{input-incomplete} items in Frt.\ (missing option fields; see Table~\ref{tab:case_table}), so interpretation of Frt.\ should be paired with stricter data-quality checks.

\subsection{Axis-Level Analysis}
\subsubsection{Process Axis (X)}
All models achieve strong average performance on the Process axis (84.4--91.9). Among the sub-benchmarks, \textbf{CIPS} is consistently high (93.2--95.3). \textbf{CSCP} is now fully covered across the evaluated models, and axis scores are reported as macro-averages over the three sub-benchmarks (CIPS/CSCP/SCMP).

\subsubsection{Variety Axis (Y)}
The Variety axis is the primary bottleneck. While \textbf{Soybeans} and \textbf{Iron Ore} are moderate (roughly mid-60s to mid-70s), performance drops sharply on \textbf{Freight Agreements (Frt.)} (28.9--48.2). This gap suggests that executable judgments under commodity-specific institutional rules and feasibility constraints remain challenging and limit practical deployment. We also flag a small number of input-incomplete items in Frt.\ (Table~\ref{tab:case_table}), motivating stronger quality controls for future releases.

\subsubsection{Cognition Axis (Z)}
Cognition performance is generally strong across models on the available sub-benchmarks (94.4--96.8 on Z-Avg). In this paper, we do not further break down localized variations on individual course-style slices.

\subsection{Case Studies and Error Taxonomy}
To make failure modes actionable, we summarize three recurring categories of \emph{executable-reasoning breaks} based on evaluation outputs and qualitative inspection: (i) \textbf{rule misreading / threshold confusion}; (ii) \textbf{feasibility or boundary-condition misjudgment}; and (iii) \textbf{ignoring financial/risk-mechanism constraints}. Table~\ref{tab:case_table} provides representative cases (plus one data-quality flag for Frt.).

\section{Conclusion}
We introduce \textbf{CSCBench} and the \textbf{PVC (Process--Variety--Cognition) 3D diagnostic framework} to evaluate LLM competence in commodity supply chains under a unified single-choice protocol, with axis/sub-benchmark reporting designed for actionable diagnostics.

Across representative LLMs under direct prompting, we observe a deployment-relevant pattern: performance remains strong on process- and course-style slices, but degrades substantially on variety-specific tasks governed by institutional rules and feasibility constraints (Table~\ref{tab:axis_results_outputs}). This suggests that, for industrial adoption, the core risk is not answer fluency but whether a model can produce \emph{executable} judgments that remain consistent with contract clauses, delivery feasibility, and (when applicable) pricing and risk-management mechanisms.

CSCBench is intended to support \emph{pre-deployment acceptance} and \emph{post-deployment regression} via PVC slices: Process helps localize failures by workflow stage, Variety exposes rule/boundary errors, and curated cases (Table~\ref{tab:case_table}) facilitate audit and remediation. Future work will strengthen data-quality controls and extend evaluation toward workflow-like tasks (e.g., clarification turns and stepwise procedures) to improve external validity in real deployments.

\section*{Limitations}
CSCBench evaluates language-based reasoning in commodity supply chains; it is not an end-to-end trading or execution system. Our current release is \textbf{text-only} and largely \textbf{single-turn}, and focuses on an objective single-choice protocol. We also identified a small number of input-incomplete items in the Freight Agreements sub-benchmark (Table~\ref{tab:case_table}), motivating stricter option-completeness checks in future releases. In addition, due to occasional credit-related API failures, some model--sub-benchmark runs may be missing and are excluded from aggregation. Extending CSCBench to interactive, multi-turn settings and incorporating time-series simulation and multimodal inputs (e.g., freight curves and charts) are important directions for future work.

\section*{Data Availability}
We will release CSCBench data, the PVC schema, prompts, and evaluation scripts, together with documentation to reproduce the axis/sub-benchmark results reported in this paper. For anonymous review we will provide an anonymized link; for the final version we will provide a permanent public repository link.

\section*{Ethics Statement}
CSCBench is constructed from publicly available rulebooks, professional curricula, and institutional reports. We do not include personal data. Since evaluation datasets can be used for test-specific optimization, we encourage reporting results by axis/sub-benchmark (not only an overall score) and conducting qualitative audits of failure cases to discourage metric overfitting and to surface unsafe overconfidence in high-stakes settings.

\bibliography{custom}

@inproceedings{hendrycks_measuring_2020,
	title = {Measuring {Massive} {Multitask} {Language} {Understanding}},
	url = {https://openreview.net/forum?id=d7KBjmI3GmQ},
	language = {en},
	urldate = {2025-12-11},
	author = {Hendrycks, Dan and Burns, Collin and Basart, Steven and Zou, Andy and Mazeika, Mantas and Song, Dawn and Steinhardt, Jacob},
	month = oct,
	year = {2020},
}

@article{cooper_supply_1997,
  title = {Supply Chain Management: More Than a New Name for Logistics},
  author = {Cooper, Martha C. and Lambert, Douglas M. and Pagh, Janus D.},
  journal = {The International Journal of Logistics Management},
  volume = {8},
  number = {1},
  pages = {1--14},
  year = {1997}
}

@article{huang_c-eval_2023,
	title = {C-{Eval}: {A} {Multi}-{Level} {Multi}-{Discipline} {Chinese} {Evaluation} {Suite} for {Foundation} {Models}},
	volume = {36},
	shorttitle = {C-{Eval}},
	url = {https://papers.nips.cc/paper_files/paper/2023/hash/c6ec1844bec96d6d32ae95ae694e23d8-Abstract-Datasets_and_Benchmarks.html},
	language = {en},
	urldate = {2025-12-11},
	journal = {Advances in Neural Information Processing Systems},
	author = {Huang, Yuzhen and Bai, Yuzhuo and Zhu, Zhihao and Zhang, Junlei and Zhang, Jinghan and Su, Tangjun and Liu, Junteng and Lv, Chuancheng and Zhang, Yikai and Lei, Jiayi and Fu, Yao and Sun, Maosong and He, Junxian},
	month = dec,
	year = {2023},
	pages = {62991--63010},
}

@article{cai_medbench_2024,
	title = {{MedBench}: {A} {Large}-{Scale} {Chinese} {Benchmark} for {Evaluating} {Medical} {Large} {Language} {Models}},
	volume = {38},
	copyright = {Copyright (c) 2024 Association for the Advancement of Artificial Intelligence},
	issn = {2374-3468},
	shorttitle = {{MedBench}},
	url = {https://ojs.aaai.org/index.php/AAAI/article/view/29723},
	doi = {10.1609/aaai.v38i16.29723},
	language = {en},
	number = {16},
	urldate = {2025-12-11},
	journal = {Proceedings of the AAAI Conference on Artificial Intelligence},
	author = {Cai, Yan and Wang, Linlin and Wang, Ye and Melo, Gerard de and Zhang, Ya and Wang, Yanfeng and He, Liang},
	month = mar,
	year = {2024},
	pages = {17709--17717},
}

@inproceedings{jin_pubmedqa_2019,
	address = {Hong Kong, China},
	title = {{PubMedQA}: {A} {Dataset} for {Biomedical} {Research} {Question} {Answering}},
	shorttitle = {{PubMedQA}},
	url = {https://aclanthology.org/D19-1259/},
	doi = {10.18653/v1/D19-1259},
	urldate = {2025-12-11},
	booktitle = {Proceedings of the 2019 {Conference} on {Empirical} {Methods} in {Natural} {Language} {Processing} and the 9th {International} {Joint} {Conference} on {Natural} {Language} {Processing} ({EMNLP}-{IJCNLP})},
	publisher = {Association for Computational Linguistics},
	author = {Jin, Qiao and Dhingra, Bhuwan and Liu, Zhengping and Cohen, William and Lu, Xinghua},
	editor = {Inui, Kentaro and Jiang, Jing and Ng, Vincent and Wan, Xiaojun},
	month = nov,
	year = {2019},
	pages = {2567--2577},
}

@inproceedings{fei_lawbench_2024,
	address = {Miami, Florida, USA},
	title = {{LawBench}: {Benchmarking} {Legal} {Knowledge} of {Large} {Language} {Models}},
	shorttitle = {{LawBench}},
	url = {https://aclanthology.org/2024.emnlp-main.452/},
	doi = {10.18653/v1/2024.emnlp-main.452},
	urldate = {2025-12-11},
	booktitle = {Proceedings of the 2024 {Conference} on {Empirical} {Methods} in {Natural} {Language} {Processing}},
	publisher = {Association for Computational Linguistics},
	author = {Fei, Zhiwei and Shen, Xiaoyu and Zhu, Dawei and Zhou, Fengzhe and Han, Zhuo and Huang, Alan and Zhang, Songyang and Chen, Kai and Yin, Zhixin and Shen, Zongwen and Ge, Jidong and Ng, Vincent},
	editor = {Al-Onaizan, Yaser and Bansal, Mohit and Chen, Yun-Nung},
	month = nov,
	year = {2024},
	pages = {7933--7962},
}

@article{guha_legalbench_2023,
	title = {{LegalBench}: {A} {Collaboratively} {Built} {Benchmark} for {Measuring} {Legal} {Reasoning} in {Large} {Language} {Models}},
	volume = {36},
	shorttitle = {{LegalBench}},
	url = {https://proceedings.neurips.cc//paper_files/paper/2023/hash/89e44582fd28ddfea1ea4dcb0ebbf4b0-Abstract-Datasets_and_Benchmarks.html},
	language = {en},
	urldate = {2025-12-11},
	journal = {Advances in Neural Information Processing Systems},
	author = {Guha, Neel and Nyarko, Julian and Ho, Daniel and Ré, Christopher and Chilton, Adam and K, Aditya and Chohlas-Wood, Alex and Peters, Austin and Waldon, Brandon and Rockmore, Daniel and Zambrano, Diego and Talisman, Dmitry and Hoque, Enam and Surani, Faiz and Fagan, Frank and Sarfaty, Galit and Dickinson, Gregory and Porat, Haggai and Hegland, Jason and Wu, Jessica and Nudell, Joe and Niklaus, Joel and Nay, John and Choi, Jonathan and Tobia, Kevin and Hagan, Margaret and Ma, Megan and Livermore, Michael and Rasumov-Rahe, Nikon and Holzenberger, Nils and Kolt, Noam and Henderson, Peter and Rehaag, Sean and Goel, Sharad and Gao, Shang and Williams, Spencer and Gandhi, Sunny and Zur, Tom and Iyer, Varun and Li, Zehua},
	month = dec,
	year = {2023},
	pages = {44123--44279},
}

@inproceedings{guo_fineval_2025,
	address = {Albuquerque, New Mexico},
	title = {{FinEval}: {A} {Chinese} {Financial} {Domain} {Knowledge} {Evaluation} {Benchmark} for {Large} {Language} {Models}},
	isbn = {979-8-89176-189-6},
	shorttitle = {{FinEval}},
	url = {https://aclanthology.org/2025.naacl-long.318/},
	doi = {10.18653/v1/2025.naacl-long.318},
	urldate = {2025-12-11},
	booktitle = {Proceedings of the 2025 {Conference} of the {Nations} of the {Americas} {Chapter} of the {Association} for {Computational} {Linguistics}: {Human} {Language} {Technologies} ({Volume} 1: {Long} {Papers})},
	publisher = {Association for Computational Linguistics},
	author = {Guo, Xin and Xia, Haotian and Liu, Zhaowei and Cao, Hanyang and Yang, Zhi and Liu, Zhiqiang and Wang, Sizhe and Niu, Jinyi and Wang, Chuqi and Wang, Yanhui and Liang, Xiaolong and Huang, Xiaoming and Zhu, Bing and Wei, Zhongyu and Chen, Yun and Shen, Weining and Zhang, Liwen},
	editor = {Chiruzzo, Luis and Ritter, Alan and Wang, Lu},
	month = apr,
	year = {2025},
	pages = {6258--6292},
}

@inproceedings{krumdick_bizbench_2024,
	address = {Bangkok, Thailand},
	title = {{BizBench}: {A} {Quantitative} {Reasoning} {Benchmark} for {Business} and {Finance}},
	shorttitle = {{BizBench}},
	url = {https://aclanthology.org/2024.acl-long.452/},
	doi = {10.18653/v1/2024.acl-long.452},
	urldate = {2025-12-11},
	booktitle = {Proceedings of the 62nd {Annual} {Meeting} of the {Association} for {Computational} {Linguistics} ({Volume} 1: {Long} {Papers})},
	publisher = {Association for Computational Linguistics},
	author = {Krumdick, Michael and Koncel-Kedziorski, Rik and Lai, Viet Dac and Reddy, Varshini and Lovering, Charles and Tanner, Chris},
	editor = {Ku, Lun-Wei and Martins, Andre and Srikumar, Vivek},
	month = aug,
	year = {2024},
	pages = {8309--8332},
}

@article{simchi-levi_managing_2004,
	title = {Managing the supply chain: the definitive guide for the business professional},
	shorttitle = {Managing the supply chain},
	url = {https://digital.library.tu.ac.th/tu_dc/frontend/Info/item/dc:272468},
	language = {eng},
	urldate = {2025-12-11},
	author = {Simchi-Levi, David and Kaminsky, Philip and Simchi-Levi, Edith},
	year = {2004},
	note = {Accepted: 2012-06-17T01:51:09Z
ISBN: 9780071435871
Number: 272468
Publisher: McGraw-Hill},
}

@book{noauthor_scor_2010,
	address = {Place of publication not identified},
	title = {{SCOR} supply chain operations reference model},
	isbn = {978-0-615-20259-4},
	language = {en},
	publisher = {The Supply Chain Council},
	year = {2010},
	note = {OCLC: 862952805},
}

@book{anderson_taxonomy_2001,
	title = {A taxonomy for learning, teaching, and assessing : a revision of {Bloom}'s taxonomy of educational objectives : complete edition},
	isbn = {978-0-321-08405-7},
	shorttitle = {A taxonomy for learning, teaching, and assessing},
	url = {https://eduq.info/xmlui/handle/11515/18824},
	language = {eng},
	urldate = {2025-12-11},
	publisher = {Addison Wesley Longman, Inc.},
	author = {Anderson, Lorin W. and Krathwohl, David R.},
	year = {2001},
}

@inproceedings{wang_scibench_2024,
	title = {{SciBench}: {Evaluating} {College}-{Level} {Scientific} {Problem}-{Solving} {Abilities} of {Large} {Language} {Models}},
	shorttitle = {{SciBench}},
	url = {https://openreview.net/forum?id=bq1JEgioLr},
	language = {en},
	urldate = {2025-12-11},
	author = {Wang, Xiaoxuan and Hu, Ziniu and Lu, Pan and Zhu, Yanqiao and Zhang, Jieyu and Subramaniam, Satyen and Loomba, Arjun R. and Zhang, Shichang and Sun, Yizhou and Wang, Wei},
	month = jun,
	year = {2024},
}

@misc{cobbe_training_2021,
	title = {Training {Verifiers} to {Solve} {Math} {Word} {Problems}},
	url = {http://arxiv.org/abs/2110.14168},
	doi = {10.48550/arXiv.2110.14168},
	urldate = {2025-12-11},
	publisher = {arXiv},
	author = {Cobbe, Karl and Kosaraju, Vineet and Bavarian, Mohammad and Chen, Mark and Jun, Heewoo and Kaiser, Lukasz and Plappert, Matthias and Tworek, Jerry and Hilton, Jacob and Nakano, Reiichiro and Hesse, Christopher and Schulman, John},
	month = nov,
	year = {2021},
	note = {arXiv:2110.14168 [cs]},
}

@inproceedings{suzgun_challenging_2023,
	address = {Toronto, Canada},
	title = {Challenging {BIG}-{Bench} {Tasks} and {Whether} {Chain}-of-{Thought} {Can} {Solve} {Them}},
	url = {https://aclanthology.org/2023.findings-acl.824/},
	doi = {10.18653/v1/2023.findings-acl.824},
	urldate = {2025-12-11},
	booktitle = {Findings of the {Association} for {Computational} {Linguistics}: {ACL} 2023},
	publisher = {Association for Computational Linguistics},
	author = {Suzgun, Mirac and Scales, Nathan and Schärli, Nathanael and Gehrmann, Sebastian and Tay, Yi and Chung, Hyung Won and Chowdhery, Aakanksha and Le, Quoc and Chi, Ed and Zhou, Denny and Wei, Jason},
	editor = {Rogers, Anna and Boyd-Graber, Jordan and Okazaki, Naoaki},
	month = jul,
	year = {2023},
	pages = {13003--13051},
}

@article{geva_strategyqa_2021,
  title = {Did Aristotle Use a Laptop? {A} Question Answering Benchmark with Implicit Reasoning Strategies},
  author = {Geva, Mor and Khot, Tushar and Segal, Elad and Shwartz, Vered and Berant, Jonathan},
  journal = {Transactions of the Association for Computational Linguistics},
  year = {2021},
  url = {https://aclanthology.org/2021.tacl-1.63/},
  doi = {10.1162/tacl_a_00465}
}

@misc{liang2023holisticevaluationlanguagemodels,
      title={Holistic Evaluation of Language Models}, 
      author={Percy Liang and Rishi Bommasani and Tony Lee and Dimitris Tsipras and Dilara Soylu and Michihiro Yasunaga and Yian Zhang and Deepak Narayanan and Yuhuai Wu and Ananya Kumar and Benjamin Newman and Binhang Yuan and Bobby Yan and Ce Zhang and Christian Cosgrove and Christopher D. Manning and Christopher Ré and Diana Acosta-Navas and Drew A. Hudson and Eric Zelikman and Esin Durmus and Faisal Ladhak and Frieda Rong and Hongyu Ren and Huaxiu Yao and Jue Wang and Keshav Santhanam and Laurel Orr and Lucia Zheng and Mert Yuksekgonul and Mirac Suzgun and Nathan Kim and Neel Guha and Niladri Chatterji and Omar Khattab and Peter Henderson and Qian Huang and Ryan Chi and Sang Michael Xie and Shibani Santurkar and Surya Ganguli and Tatsunori Hashimoto and Thomas Icard and Tianyi Zhang and Vishrav Chaudhary and William Wang and Xuechen Li and Yifan Mai and Yuhui Zhang and Yuta Koreeda},
      year={2023},
      eprint={2211.09110},
      archivePrefix={arXiv},
      primaryClass={cs.CL},
      url={https://arxiv.org/abs/2211.09110}, 
}

@misc{srivastava2023imitationgamequantifyingextrapolating,
      title={Beyond the Imitation Game: Quantifying and extrapolating the capabilities of language models}, 
      author={Aarohi Srivastava and Abhinav Rastogi and Abhishek Rao and Abu Awal Md Shoeb and Abubakar Abid and Adam Fisch and Adam R. Brown and Adam Santoro and Aditya Gupta and Adrià Garriga-Alonso and Agnieszka Kluska and Aitor Lewkowycz and Akshat Agarwal and Alethea Power and Alex Ray and Alex Warstadt and Alexander W. Kocurek and Ali Safaya and Ali Tazarv and Alice Xiang and Alicia Parrish and Allen Nie and Aman Hussain and Amanda Askell and Amanda Dsouza and Ambrose Slone and Ameet Rahane and Anantharaman S. Iyer and Anders Andreassen and Andrea Madotto and Andrea Santilli and Andreas Stuhlmüller and Andrew Dai and Andrew La and Andrew Lampinen and Andy Zou and Angela Jiang and Angelica Chen and Anh Vuong and Animesh Gupta and Anna Gottardi and Antonio Norelli and Anu Venkatesh and Arash Gholamidavoodi and Arfa Tabassum and Arul Menezes and Arun Kirubarajan and Asher Mullokandov and Ashish Sabharwal and Austin Herrick and Avia Efrat and Aykut Erdem and Ayla Karakaş and B. Ryan Roberts and Bao Sheng Loe and Barry Bond and Basil Mustafa and Ben Swirsky and Benji Smith and Benjamin Steinhardt and Benoît Kessler and Bhargav Kanuparthi and Bhaskar Mitra and Bhavana Dalvi and Bhuvana Ramabhadran and Bianca Sap and Binhang Yuan and Blake Wulfe and Bohan Zhang and Brian Hutchinson and Brian Lin and Brooks R. H. S. S. E. and Bruna da Silva and Caiming Xiong and Calum McConnell and Cameron Clark and Canwen Xu and Carissa Schoenick and Casey Meng and Catherine D’Ignazio and Ce Zhang and Chaitanya Parashar and Chandan Singh and Charles Chen and Charlie Nash and Chen Zhao and Chengtao Li and Chia-Wei Hsu and Chin-Yew Lin and Chinmay Hegde and Chris Paxton and Chris Thomson and Christian Bjorndahl and Christian H. M. Wilson and Christopher D. Manning and Christopher Ré and Christopher W. Choi and Clara Vania and Clark Barrett and Colin Raffel and Conor O’Donnell and Courtney Napoles and Daniel Levy and Daniel Scales and Daniel Ziegler and Danny Hernandez and Danyal M. and Daphne Ippolito and Dara Bahri and David Berthelot and David Chiang and David Jurgens and David Krueger and David Kroon and David L. and Deep Ganguli and Deepak Narayanan and Dennis Chen and Deniz Yuret and Devi Parikh and Dhruv Mahajan and Diana Acosta-Navas and Diederik P. Kingma and Dmitry Krotov and Dmitry Lepikhin and Dominik Schmid and Don Metzler and Douwe Kiela and Drew A. Hudson and Eda Kazak and Edoardo Maria Ponti and Edouard Grave and Edwin Simpson and Ehsan Abdalrahman and Ekaterina Krivosheev and Elad Segal and Eleanor Birrell and Elena Black and Elena Voita and Elliotte Rusty Harold and Emily Dinan and Emily Mueller and Enrico Santus and Ethan Perez and Eugene Yang and Eunsol Choi and Faisal Ladhak and Faraz Hussain and Felicity and Freda Rong and Gabe Grand and Gabriel Ilharco and Gaurav and Ge Gao and Geet and Geoffrey Hinton and George D. and Giorgi and Greg Brockman and Hailey Schoelkopf and Hao Zhang and Hao Zheng and Harrison and He He and Hila and Holger Schwenk and Hongyu Ren and Hrishikesh and Hui and Ian Osband and Ian Tenney and Ilya Sutskever and Ishaan Gulrajani and Ivan and Jaehoon Lee and Jakob and James and Jan and Jason Wei and Javier and Jeff and Jennifer and Jeremy and Jessica and Jian and Jinhua and Jim and Jin and Jinyu and Joao and John and Jonathan and Joseph and Josh and Jue Wang and Julia and Julien and Jun and Junxian He and Karthik and Keshav Santhanam and Kevin and Kh and Kiat and Kilian and Kim and Kory and Krish and Krishna and Kyla and Kyle and L. and Lao and Laura and Laurel Orr and Leila and Lianmin Zheng and Lila and Lin and Linda and Lisa and Lucia Zheng and Luke and M. and Machel and Madeline and Maggie and Mahdi and Maithra and Mallory and Manoj and Marco and Marie and Mark and Martin and Mary and Matt and Matthew and Max and May and Megan and Mehul and Mert Yuksekgonul and Michael and Michihiro Yasunaga and Mike and Milad and Mirac Suzgun and Mohammad and Mohit and Molly and Moran and N. and Nate and Nathan Kim and Neel Guha and Neil and Nicholas and Nick and Nicole and Niladri Chatterji and Nir and Noam and Noah and Nour and O. and Omar Khattab and Oriol and Owen and P. and Paulius and Percy Liang and Peter Henderson and Petr and Philip and Pranav and Qian Huang and R. and Rishi Bommasani and Robert and Rodrigo and Rohit and Ryan Chi and S. and Sagnik and Sahil and Sang Michael Xie and Sarah and Sasha and Sebastian and Shibani Santurkar and Shreya and Siva and So and Sophie and Srini and Surya Ganguli and Szymon and T. and Tatsunori Hashimoto and Taylor and Thomas Icard and Tianyi Zhang and Tim and Timo and Tom and Tony Lee and Tristan and U. and V. and Victor and Vishrav Chaudhary and Will and William Wang and Wouter and X. and Xuechen Li and Y. and Yian Zhang and Yifan Mai and Yuhuai Wu and Yuhui Zhang and Yuta Koreeda and Z. and Zeming and Zico and Zi Lin and Zoubin Ghahramani},
      year={2023},
      eprint={2206.04615},
      archivePrefix={arXiv},
      primaryClass={cs.CL},
      url={https://arxiv.org/abs/2206.04615}, 
}

@misc{zheng2023judgingllmasajudgemtbenchchatbot,
      title={Judging LLM-as-a-Judge with MT-Bench and Chatbot Arena}, 
      author={Lianmin Zheng and Wei-Lin Chiang and Ying Sheng and Siyuan Zhuang and Zhanghao Wu and Yonghao Zhuang and Zi Lin and Zhuohan Li and Dacheng Li and Eric P. Xing and Hao Zhang and Joseph E. Gonzalez and Ion Stoica},
      year={2023},
      eprint={2306.05685},
      archivePrefix={arXiv},
      primaryClass={cs.CL},
      url={https://arxiv.org/abs/2306.05685}, 
}

@misc{chiang2024chatbotarenaopenplatform,
      title={Chatbot Arena: An Open Platform for Evaluating LLMs by Human Preference}, 
      author={Wei-Lin Chiang and Lianmin Zheng and Ying Sheng and Anastasios Nikolas Angelopoulos and Tianle Li and Dacheng Li and Hao Zhang and Banghua Zhu and Michael Jordan and Joseph E. Gonzalez and Ion Stoica},
      year={2024},
      eprint={2403.04132},
      archivePrefix={arXiv},
      primaryClass={cs.AI},
      url={https://arxiv.org/abs/2403.04132}, 
}

@misc{hendrycks2021cuadexpertannotatednlpdataset,
      title={CUAD: An Expert-Annotated NLP Dataset for Legal Contract Review}, 
      author={Dan Hendrycks and Collin Burns and Anya Chen and Spencer Ball},
      year={2021},
      eprint={2103.06268},
      archivePrefix={arXiv},
      primaryClass={cs.CL},
      url={https://arxiv.org/abs/2103.06268}, 
}

@inproceedings{chalkidis-etal-2022-lexglue,
    title = "{L}ex{GLUE}: A Benchmark Dataset for Legal Language Understanding in {E}nglish",
    author = "Chalkidis, Ilias  and
      Jana, Abhik  and
      Hartung, Dirk  and
      Bommarito, Michael  and
      Androutsopoulos, Ion  and
      Katz, Daniel  and
      Aletras, Nikolaos",
    editor = "Muresan, Smaranda  and
      Nakov, Preslav  and
      Villavicencio, Aline",
    booktitle = "Proceedings of the 60th Annual Meeting of the Association for Computational Linguistics (Volume 1: Long Papers)",
    month = may,
    year = "2022",
    address = "Dublin, Ireland",
    publisher = "Association for Computational Linguistics",
    url = "https://aclanthology.org/2022.acl-long.297/",
    doi = "10.18653/v1/2022.acl-long.297",
    pages = "4310--4330"
}

@misc{koreeda2021contractnlidatasetdocumentlevelnatural,
      title={ContractNLI: A Dataset for Document-level Natural Language Inference for Contracts}, 
      author={Yuta Koreeda and Christopher D. Manning},
      year={2021},
      eprint={2110.01799},
      archivePrefix={arXiv},
      primaryClass={cs.CL},
      url={https://arxiv.org/abs/2110.01799}, 
}

@misc{chen2022finqadatasetnumericalreasoning,
      title={FinQA: A Dataset of Numerical Reasoning over Financial Data}, 
      author={Zhiyu Chen and Wenhu Chen and Charese Smiley and Sameena Shah and Iana Borova and Dylan Langdon and Reema Moussa and Matt Beane and Ting-Hao Huang and Bryan Routledge and William Yang Wang},
      year={2022},
      eprint={2109.00122},
      archivePrefix={arXiv},
      primaryClass={cs.CL},
      url={https://arxiv.org/abs/2109.00122}, 
}

\appendix
\section{Case Table}
\label{sec:appendix_cases}

\begin{center}
  \small
  \setlength{\tabcolsep}{2pt}
  \renewcommand{\arraystretch}{1.15}
  \begin{tabularx}{\textwidth}{@{}>{\raggedright\arraybackslash}p{2.4cm} >{\raggedright\arraybackslash}X c >{\raggedright\arraybackslash}X@{}}
  \toprule
  \textbf{Slice} & \textbf{Prompt summary} & \textbf{Gold/Pred.} & \textbf{Key failure / take-away} \\
  \midrule
  Y--Soybeans (delivery rules) & Moisture upper bound for deliverable grade & B / C & \textbf{Rule confusion}: adjacent thresholds are easily conflated, yielding non-executable judgments. \\
  Y--Iron Ore (delivery regime) & Contract storage capacity vs.\ deliverability decision & B / A & \textbf{Boundary misjudgment}: misreading boundary conditions as physical capacity constraints breaks feasibility reasoning. \\
  Y--Soybeans (hedging/risk) & Hedge contract choice under liquidity vs.\ tenor match & B / C & \textbf{Risk-mechanism omission}: ignoring liquidity/tenor trade-offs distorts executable hedging decisions. \\
  X--CSCP (process systems) & Typical use of APS in an end-to-end workflow & D / C & \textbf{Concept boundary}: confusing neighboring system roles shifts process-stage decisions. \\
  Y--Frt.\ (freight agreements) & Turnaround scenario with missing option text & B / N/A & \textbf{Data quality}: missing option fields can affect Frt.\ scoring; enforce option-completeness checks in future releases. \\
  \bottomrule
  \end{tabularx}
  \captionof{table}{Representative failure cases from evaluation outputs (IDs omitted for readability). ``N/A'' indicates that the output is non-parsable under our letter-extraction rule.}
  \label{tab:case_table}
\end{center}

\end{document}